\documentclass[conference]{IEEEtran}
\IEEEoverridecommandlockouts
\usepackage{cite}
\usepackage{amsmath,amssymb,amsfonts}
\usepackage{algorithmic}
\usepackage{graphicx}
\usepackage{textcomp}
\usepackage{xcolor}
\usepackage{url}

\usepackage{multirow} 
\usepackage{ragged2e}
\usepackage{array}

\definecolor{darkgreen}{rgb}{0.0, 0.5, 0.0}

\def\BibTeX{{\rm B\kern-.05em{\sc i\kern-.025em b}\kern-.08em
    T\kern-.1667em\lower.7ex\hbox{E}\kern-.125emX}}
\begin{document}

\title{Context-Awareness and Interpretability of Rare
Occurrences for Discovery and Formalization of
Critical Failure Modes

\thanks{Approved for Public Release; Distribution Unlimited. Public Release Case Number 25-0079. 
\\©2025 The MITRE Corporation. ALL RIGHTS RESERVED.
}
}


\author{
\IEEEauthorblockN{1\textsuperscript{st} Sridevi Polavaram}
\IEEEauthorblockA{\textit{MITRE CORP.}\\
Virginia, USA \\
spolavaram@mitre.org}
\and
\IEEEauthorblockN{2\textsuperscript{nd} Xin Zhou}
\IEEEauthorblockA{\textit{MITRE CORP.}\\
Austin, USA \\
xzhou@mitre.org}
\and
\IEEEauthorblockN{3\textsuperscript{rd} Meenu Ravi}
\IEEEauthorblockA{\textit{MITRE CORP.}\\
Virginia, USA\\
mravi@mitre.org}
\and
\IEEEauthorblockN{4\textsuperscript{th} Mohammad Zarei}
\IEEEauthorblockA{\textit{MITRE CORP.}\\
San Diego, USA \\
mzarei@mitre.org}
\and
\IEEEauthorblockN{5\textsuperscript{th} Anmol Srivastava}
\IEEEauthorblockA{
\textit{MITRE CORP.}\\
Virginia, USA \\
asrivastava@mitre.org}
}

\maketitle

\begin{abstract}

Vision systems are increasingly deployed in critical domains such as surveillance, law enforcement, and transportation. However, their vulnerabilities to rare or unforeseen scenarios pose significant safety risks. To address these challenges, we introduce Context-Awareness and Interpretability of Rare Occurrences (CAIRO), an ontology-based human-assistive discovery framework for failure cases (or CP - Critical Phenomena) detection and formalization. CAIRO by design incentivizes human-in-the-loop for testing and evaluation of criticality that arises from misdetections, adversarial attacks, and hallucinations in AI black-box models. Our robust analysis of object detection model(s) failures in automated driving systems (ADS) showcases scalable and interpretable ways of formalizing the observed gaps between camera perception and real-world contexts, resulting in test cases stored as explicit knowledge graphs (in OWL/XML format) amenable for sharing, downstream analysis, logical reasoning, and accountability.

\end{abstract}


\section{Introduction}
Formal verification techniques are a norm in chip design, but they remain elusive in computer vision (CV) applications. The reason being CV applications are deemed open-ended, often trained on millions of data and billions of parameters to learn a few hundreds of labels. Finetuning practices are commonly used to tailor them to specific needs, but with no standard testing procedures in place providing guidance for their application to ensure fail-safe behaviors, critical systems like Autonomous Vehicles (AV) are bound to fail~\cite{reuters_cruise_2024}. Here, we bring forth two crucial missing factors in testing and evaluation of critical systems: (a) The inability to have a meaningful mapping between the perception of an AV and its human counterpart and (b) the oversight of the CV models to human-factors like fairness, trustworthiness, and collaboration.

While the large data AI models are popular for their open world assumption (OWA) to learn new concepts, our goal is to design CAIRO for mitigating the disparities between human cognition and AV perception. For this, we bring in schema-based closed world approach (CWA), opting for a middle-ground hybrid OWA-CWA~\cite{Bergman2016TheOW} solution to enable human-machine teaming and leveraging domain expertise. 

\begin{figure}
\centering
\includegraphics[width=\linewidth]{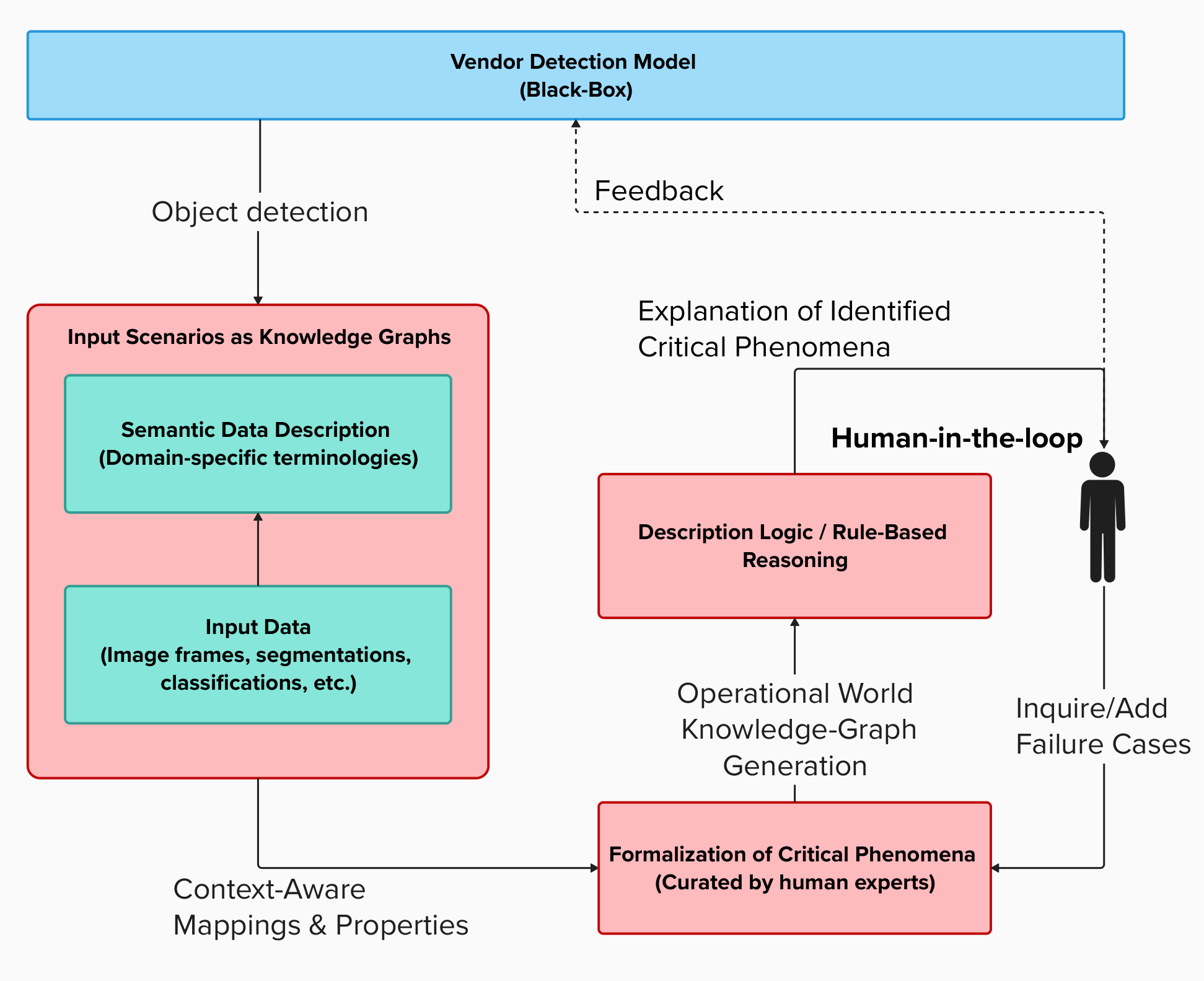}
\caption{Testing \& evaluation (T\&E) with Knowledge Graphs (KGs) in green and human-in-the-loop (HITL) interpreting detection model inferences in blue through CAIRO processes in red.}
\label{cairo_intro}
\end{figure}
As shown in Figure~\ref{cairo_intro}, during T\&E, the key is to impose finiteness for effectively aligning system behavior with human perception by evaluating the models based on features that humans consider as relevant. To achieve this, we propose two levels of feature control: (a) the use of standardized terminologies as nodes \& edges in the KGs and (b) including HITL expertise in CP rule formulation. Our approach exemplifies successful management of valid combinations of known failure cases, even allowing for systematic validation of CV models by loading the KGs modularly to fit the scope of the use case. 

To the best of our knowledge CAIRO is the first of its kind to adopt ontology-based grounding for analyzing and evaluating CV model performance. We extend our experiments to exemplify the robustness of CP formalization in addressing adversarial criticality as well. Providing the benefit of graph-based analysis despite the attack success from physically realizable patch attacks. The mutually competing nature of operational relevance vs KG provenance plays a powerful role in human-centered verification \& validation (V\&V) of critical systems.

%
\section{Related works}
This section reviews the foundational concepts and related works that inform the development of the CAIRO framework. It begins with an introduction to KGs as tools for enhancing AI interpretability. Next, it examines their applications in autonomous systems, focusing on perception, navigation, and scenario modeling. Finally, it identifies gaps in current T\&E approaches.

\subsection{Background}
KGs epitomize comprehensible artificial intelligence and serve as an essential process for making AI models human-understandable~\cite{CAI}. They are used to abstract complex thought
patterns~\cite{besta2024graph}, convert bodies of work into interconnected concepts ~\cite{rahulnyk_knowledge_graph}, integrate data, and provide collaborative open-sourced ecosystems for analytic and search capabilities ~\cite{Polavaram2017, putman2024monarch} in many domains. 

In the domain of AV, ontologies have been significant in processing sensor data to support path navigation and mission planning based on environmental perception~\cite{kammel2009team}, as demonstrated in the 2007 DARPA Urban Challenge featuring driverless vehicles~\cite{darpa_urban_challenge2009}. Ontologies establish a common language for modeling the operational world. However, creating robust scenarios for keeping the ontological models up-to-date during comprehensive safety analysis and testing remains a challenge~\cite{bagschik2017ontology}. The Society of Automotive Engineers (SAE~\cite{SAE}) consistently provide standards for relevant ontologies and lexicons. The Operational World Model Ontology for ADS part 1~\cite{Czarnecki} \& part 2~\cite{czarnecki2018operational} provided a well-defined scope for operational design domain (ODD). More recently, the ODD has been digitized, as Automotive Urban Traffic Ontology (A.U.T.O)~\cite{scholtes2021sixlayer}, incorporating the six-layer ODD model (6LM) architecture into the operational world schema. The availability of NHTSA test cases~\cite{Thorn2018AFF}, along with the ability to represent scenarios in the 6LM format~\cite{scholtes2022omegaformat}, and the manually designed CP modeling approaches~\cite{westhofen2022using} together provide a complete template to design safety critical use cases for ADS.

The success of CAIRO will further the automation of knowledge graph updates for test cases, fostering a community for knowledge sharing, similar to efforts in the life sciences, such as Bioportal~\cite{whetzel2011bioportal} and Monarch Initiative~\cite{putman2024monarch}. This research has the potential to revolutionize V\&V communication scalability by allowing to condense and amplify the impact of published content like e.g., accident reports on AV vs. human driver comparisons, enabling AVs to operate more safely and efficiently, as envisioned by NIST.

\subsection{Terminologies - Notations - Definitions}

In this paper, failure cases or edge cases are referred to as CP, following the AV V\&V of safety principles~\cite{neurohr2021criticality} that expand upon the European AV projects PEGASUS~\cite{pegasus} and ENABLE-S3~\cite{trimis}, both of which adopt scenario-based approaches at SAE Level 3. A CP is defined as ``A single influencing factor, or a combination thereof, that is associated with increased criticality in a scene or scenario”~\cite{westhofen2022using}.

 As defined in Westhofen et al.~\cite{westhofen2022using} a triple \( N = (N_R, N_C, N_I) \) is a set of role names \( N_R \), concept names \( N_C \), and individual names \( N_I \). The T-Box (Terminology-Box) is a finite set of terminological assertions of general concepts such as \( C \sqsubseteq C' \) for two concept names \( C, C' \in N_C \), as well as role inclusion axioms such as \( r \sqsubseteq r' \) for two role names \( r, r' \in N_R \). The A-Box (Assertion-Box) is a finite set of concept and role assertions which bind to known individuals that are detected by the detection model (DM), such as \( C(x) \) and \( r(x, x') \) for \( x, x' \in N_I \), \( C \in N_C \), and \( r \in N_R \). 

The logical operations like negation (\( \neg \)), intersection (\( \cap \)), union (\( \cup \)), and universal (\( \forall r.C \)) and existential (\( \exists r.C \)) role quantification are supported by Description Logic (DL). We leverage the decisiveness of DL and the expressiveness of the Semantic Web Rule Language (SWRL) in formulating the CP as originally introduced by Antoniou et al.~\cite{antoniou}. An SWRL rule over variables \( V \) is a Horn clause of a set of antecedent rule atoms \( RA \) over \( V \) and precedent rule atoms \( RP \) over \( V' \subseteq V \), i.e., of the form:
\[
\bigwedge_{ra \in RA} ra \implies \bigwedge_{rp \in RP} rp.
\]

\begin{figure}[h!]
\centering
\includegraphics[width=\linewidth]{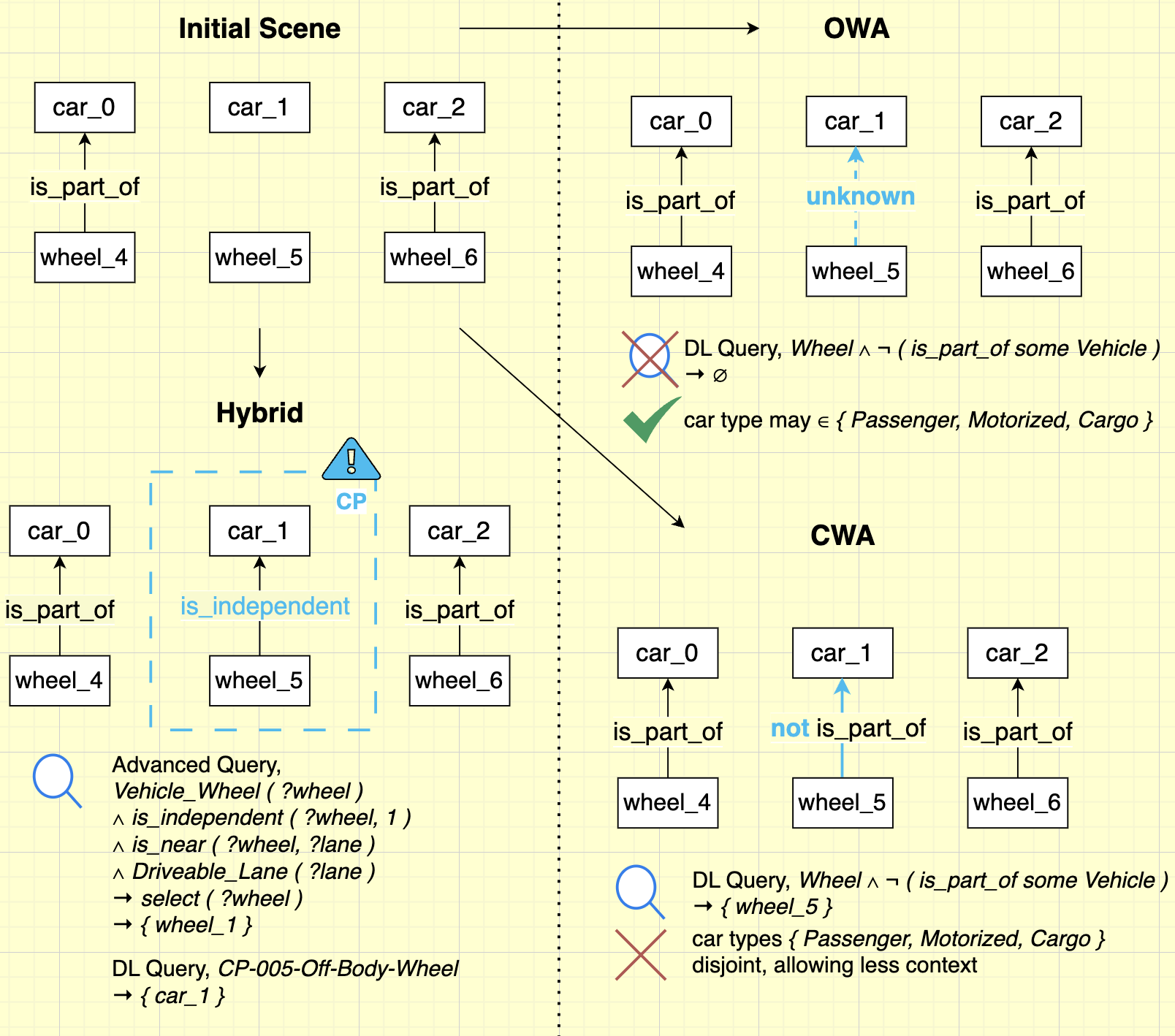}
\caption{OWA, CWA, and hybrid approach comparison in KG representation.}
\label{fig:owa_cwa_hybrid}
\end{figure}

\begin{table}[h!]
\centering
\caption{(a) Axioms of T-Box and A-box assertions for the \texttt{car} concept, (b) Rules in SWRL describing more complex behaviors, such as a missing license plate.}

\renewcommand{\arraystretch}{1.3} 

\begin{tabular}{|p{3.7cm}|p{4.3cm}|} 
\hline
\multicolumn{1}{|c|}{\textbf{(a) Axioms}} & \multicolumn{1}{c|}{\textbf{(b) SWRL Rules}} \\ \hline

\texttt{car\_1 is\_a Vehicle} & \textbf{DL Query:} \\ \hline 
\raggedright \texttt{car\_1 has no\_plate} \arraybackslash & \raggedright \texttt{passenger\_car} $\sqcap$ \\ 

$\neg$ \texttt{has\_part\\}
\texttt{(license\_plate)} \arraybackslash \\ \hline
\raggedright \texttt{car\_1 has\_part wheels} \arraybackslash & \raggedright \textbf{Advanced SWRL Query:} \arraybackslash \\ \hline
\raggedright \texttt{car\_1 has\_brake} \\
\raggedright \texttt{\_lights}  & 
\multirow{2}{*}{
\parbox[t]{5cm}{
\texttt{l4\_d:Passenger\_Car(?c)} \\
\texttt{$\land$ phys:no\_plate(?c,?p)} \\
\texttt{$\land$ swrb:equal(?p, 1)} \\
\texttt{$\land$ sqwrl:select(?c)}

} }\\ \cline{1--1}
\raggedright \texttt{car\_1 number\_of\_wheels } $\leq 4$ \arraybackslash & \arraybackslash \\ \hline

\end{tabular}
\label{tab:axioms_swr_rules}
\end{table}

\begin{figure*}[!ht]
\centering
\includegraphics[width=\textwidth]{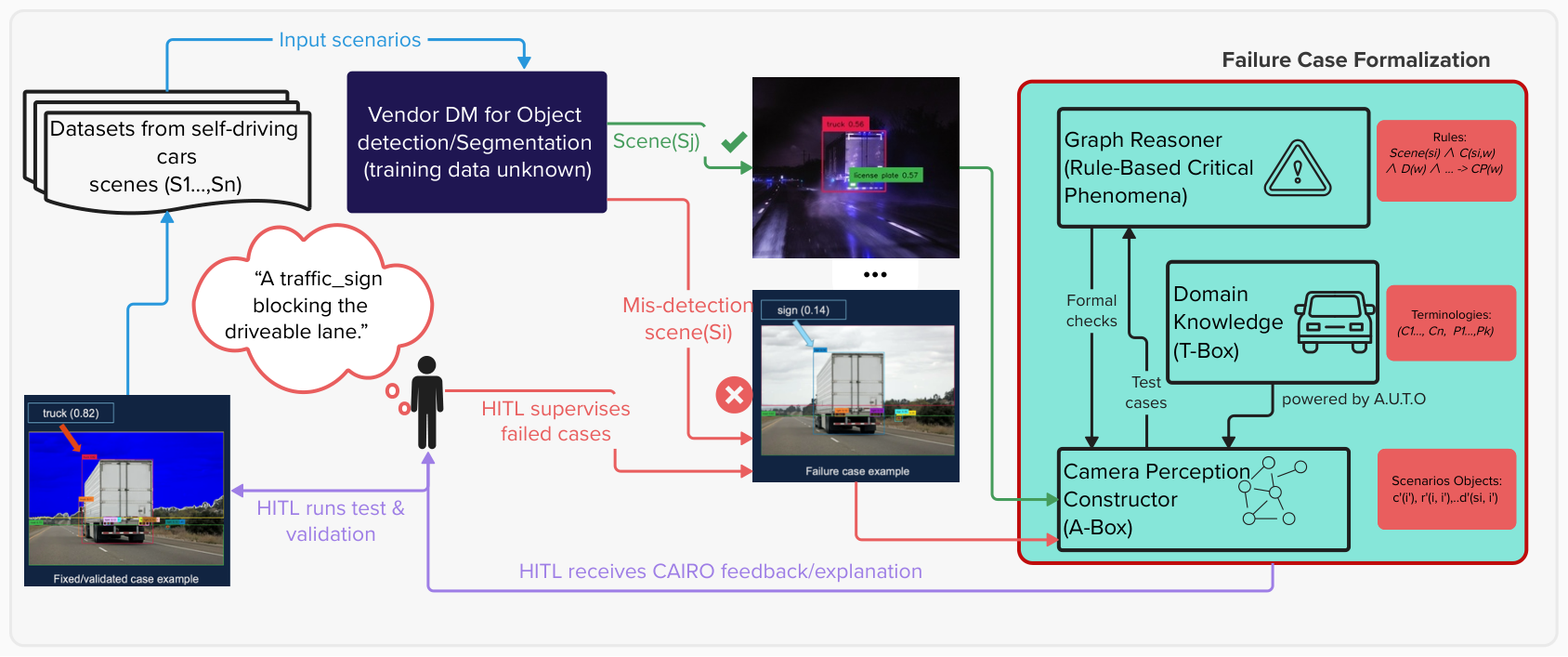}
\caption{CAIRO pipeline: A comprehensive framework integrating KGs, logical reasoning, and HITL validation to enhance the reliability and interpretability of critical phenomena detection in autonomous systems.}
\label{fig:pipeline}
\end{figure*}

Although the ODD is convenient to model as OWA, the actual combinatorial number of cases that trigger a CP are finite. The convergence of CP inherently adopts a CWA in the rules that function as "find" criteria of causality, as axioms are connected into Horn rules. Figure~\ref{fig:owa_cwa_hybrid} illustrates this concept with an input scene featuring three cars. A car as an object of interest is represented by a set of axioms in Table~\ref{tab:axioms_swr_rules} (a) that describe its physical, spatial, and temporal properties. A valid combination of axioms form a rule as shown in Table~\ref{tab:axioms_swr_rules} (b) to describe the find criteria for CP w.r.t the ego-vehicle's perception.

As indicated in Figure~\ref{fig:owa_cwa_hybrid}, under OWA, absence of a relationship does not imply that its negation is true; instead, it indicates a lack of knowledge about that relationship. As a result, a DL query fails when queried for a \texttt{wheel} that is not related to \texttt{Vehicle}, but on the flip-side OWA offers the flexibility to allow for multi-class subsumption, i.e., \texttt{car} could refer to a \texttt{passenger car}, or \texttt{motorized car}, or \texttt{cargo car}, depending on the use case. Under CWA this flexibility is lost due to strict disjointedness, but negation operations are enabled in DL queries. The hybrid approach allows for formalization, combining multiple axioms into a contextually relevant rule such as \texttt{CP-005-Off-Body-Wheel} for detecting a \texttt{wheel} found independent from its body and is \texttt{near a driveable lane}. The rule is triggered whenever the specified criteria are met, but the way this can visually manifest in an input scene can vary significantly depending on the DM and the real-world scenario.

\begin{table*}[!ht]
\caption{Examples of mixed set of behavioral CP }
\begin{center}
\renewcommand{\arraystretch}{1.5} 
\begin{tabular}{|p{0.5cm}|p{3.8cm}|p{2.5cm}|p{6cm}|p{3.5cm}|} 
\hline
\textbf{ID} & \textbf{Natural Language Query} & \textbf{Property Types} & \textbf{SWRL Query (Rules)} &  \textbf{Scene Description} \\
\hline

\texttt{CP 0001} & Find vulnerable road user occluded by at least 10\% and are wearing color gray & Dynamic object Properties & 
\parbox[t]{6.2cm}{\raggedright
\texttt{l4\_d:Vulnerable\_Road\_User(?v)}  $\land$
\texttt{perc:has\_high\_occlusion(?v, true)} 
$\land$ \texttt{phys:has\_color(?v, phvs:Gray)} 
$\rightarrow$ \texttt{sqwrl:select(?v)}} & 
Scene: A busy urban intersection showing pedestrians crossing and vehicles amidst city buildings. \\
\hline

\texttt{CP 0002} & Find strollers near a driveable lane in scene 1 but not scene 2 & Temporal relations & 
\parbox[t]{6.2cm}{\raggedright
\texttt{l4\_d:Stroller(?s)}  $\land$ \texttt{traf:traffic\_model\_element} \texttt{\_property(?s, ?scene) } $\land$ \texttt{traf:traffic\_model\_element} \texttt{\_property(?s, ?scene1)} $\land$ \texttt{differentFrom(?scene, traf:scene2)} $\rightarrow$ \texttt{sqwrl:select(?s)}} & 
Scenario: A busy urban intersection showing pedestrians crossing and vehicles amidst city buildings. \\
\hline


\texttt{CP 0003} & Find bicycle that is near pedestrians and is in a busy crosswalk & Behavioral Unpredictability, Combining CP & 
\parbox[t]{6.2cm}{\raggedright
\texttt{l4\_d:Bicycle(?b)} $\land$ 
\texttt{phys:is\_in\_proximity(?b, ?cs) } $\land$ 
\texttt{l1\_c:Crossing\_Site(?cs)}  $\land$ 
\texttt{phys:is\_in\_proximity(?cs, ?vru) } $\rightarrow$ 
\texttt{sqwrl:select(?b)}} & 
Scene: A busy urban intersection showing pedestrians crossing and vehicles amidst city buildings. \\
\hline

\texttt{CP 0004} & Find cars less than 50 distance away from driver and has missing license plate & 
\raggedright Distance~Metrics,~Behavioral~Unpredictability & 
\parbox[t]{6.2cm}{\raggedright
\texttt{l4\_d:Passenger\_Car(?car) $\land$} 
\texttt{phys:no\_plate(?car, ?p) $\land$} 
\texttt{swrb:equal(?p, 1) $\land$} 
\texttt{phys:has\_distance(?car, ?distance) $\land$} 
\texttt{swrb:lessThan(?distance, 50.0) $\rightarrow$}
\texttt{sqwrl:select(?car)}} & 
Scene: A white SUV drives along a curving desert road surrounded by bush/plants. \\
\hline

\texttt{CP 0005} & Find detatched wheel near a vehicle lane & 
\raggedright Behavioral~Unpredictability & 
\parbox[t]{6.2cm}{\raggedright
\texttt{l4\_d:Vehicle\_Wheel(?w)  $\land$} 
\texttt{phys:is\_independent(?w, 1)  $\land$} 
\texttt{phys:is\_near(?w, ?l)  $\land$} 
\texttt{l1\_c:Driveable\_Lane(?l) $\land$} 
\texttt{ $\rightarrow$ sqwrl:select(?w)} }& 
Scene: A busy urban intersection showing pedestrians crossing and vehicles amidst city buildings. \\
\hline

\end{tabular}
\label{tab:cp_queries_combined}
\end{center}
\end{table*}

\section{Methodology}

As depicted in Figure 3 our methodology consists of detection, segmentation, feature extraction, ontology graph construction, formalization, reasoning, and HITL validation. This pipeline processes images or video frames, extracts individuals, and constructs structured ontological graphs for iterative reasoning to ultimately identify and validate root causes of perception failures in AV systems. We build on well-established and validated resources that are compatible with the 6LM architecture, ensuring the OWL files generated from the experiments are consistent and are easy-to-share for further analysis and reporting of failure cases. Some of those example artifacts can be found in CAIRO\_Artifacts~\cite{Artifacts} repository.

\subsection{Detection and Segmentation}

To identify regions of interest within an image that contribute to model misclassifications or missed detections, we use a combination of three DMs: Faster R-CNN~\cite{fastrcnn}, GroundingDINO \cite{liu2023grounding}, and CLIP. Faster R-CNN and GroundingDINO are used for initial object detection, and their bounding boxes are refined and extracted using the Segment Anything Model (SAM)~\cite{kirillov2023segment}. We use CLIP~\cite{clip} to classify regions segmented by SAM. This involves cropping the areas defined by SAM's bounding boxes and having CLIP classify them. CLIP helps fill the gaps left by text-to-vision models, which may miss important details. For example, Faster R-CNN is limited by its predefined set of classes, and GroundingDINO relies heavily on text prompts. By using CLIP, we ensure a more comprehensive classification of segmented regions. Additionally, we integrated the Canny edge detector~\cite{Canny} to predict driving directions, enhancing edge-based analysis.

\subsection{Feature Extraction}

For each segmented region (node), we extract a comprehensive set of features (edges) to capture its contextually rich physical, spatial, and relational characteristics. These features include, but are not limited to:
\begin{itemize}
    \item Physical Properties: attributes like color, texture, and vehicle components.
    \item Spatial Attributes: metrics such as distance, size, relative angle, and spatial relationships (e.g., \texttt{is\_left\_of, is\_right\_of, is\_occluded\_by}).
    \item Statistical Metrics: features like \texttt{logits} and other model outputs that quantify prediction certainty.
    \item Ego-centric relationships: semantic relationships between regions, such as \texttt{has\_distance, has\_yaw, has\_part}, and so on.
\end{itemize}

\begin{figure*}[ht]
\centering
\includegraphics[width=\textwidth]{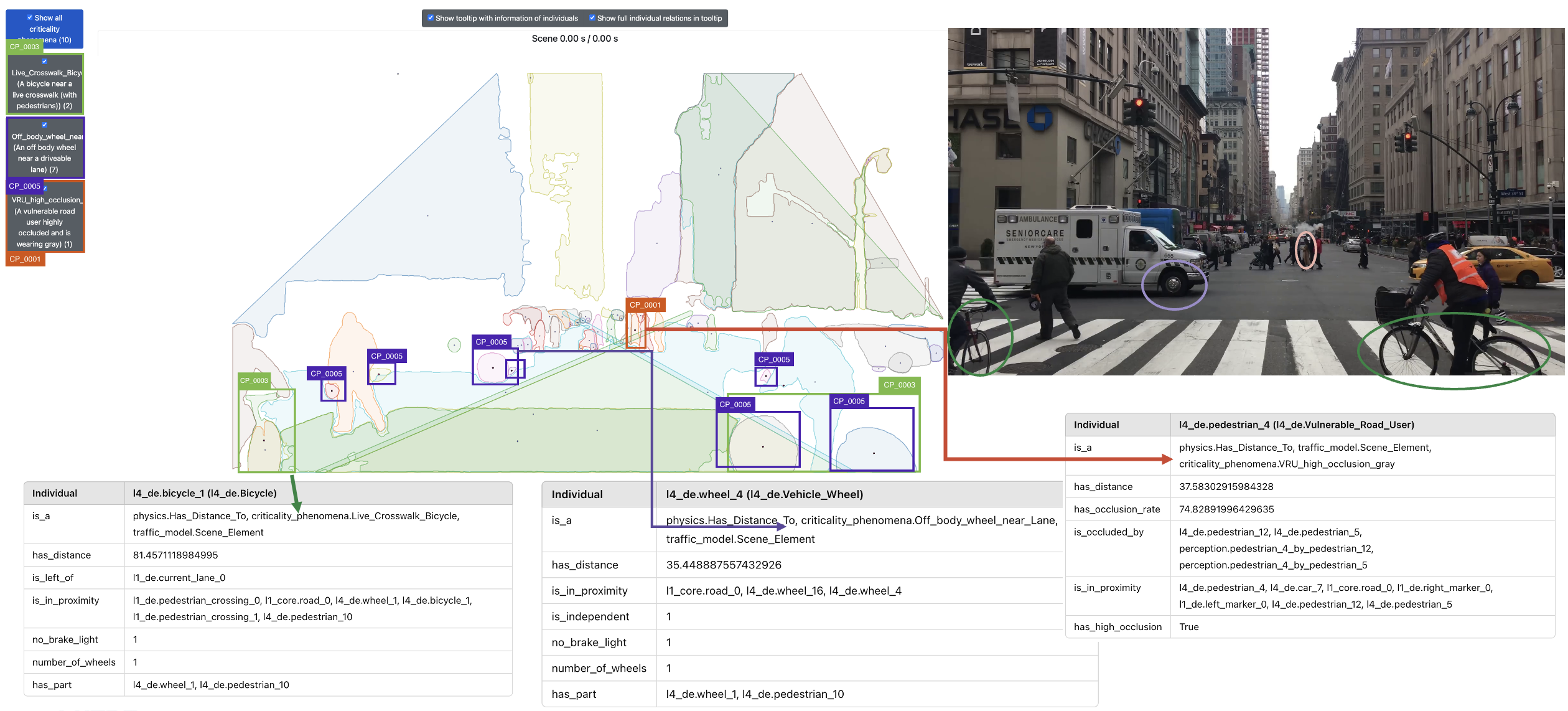}
\caption{Demonstration of a set of behavioral CP: camera mis-perception and human CP. \texttt{CP\_0003} (\textcolor{darkgreen}{green}) shows behavioral CP - illegitimate use of a crosswalk by the bicyclists. \texttt{CP\_0005} (\textcolor{violet}{purple}) finds behavioral CP - detached wheels being close to vehicle lane. \texttt{CP\_0001}(\textcolor{orange}{orange}) finds a dynamic CP - vulnerable road users that are highly occluded.}
\label{other_cps}
\end{figure*}

\subsection{Ontology Graph Construction}

 We construct a scalable directed graph that encapsulates the segmented regions with the real-world concepts. For consistent representation we leverage existing systems-level schema design like the 6LM and other published ontologies as needed for adding new concepts/properties. 
\begin{itemize} 
\item \textbf{Nodes (Concepts)}: Represent segmented regions, which are instances of defined concepts (e.g., \texttt{Vehicle}, \texttt{Pedestrian}, \texttt{TrafficLight}). Each node is annotated and gets instantiated by detected scene elements. 
\item \textbf{Edges (Relations)}: Capture the relationships between nodes, such as \texttt{is\_left\_of}, \texttt{is\_right\_of}, \texttt{is\_occluded\_by}, or context-specific properties (e.g., \texttt{has\_part}, \texttt{has\_distance}, \texttt{has\_yaw}) with a range of values. \item \textbf{Scene}: A scene corresponds to a single frame, representing a snapshot of segmented regions from an ego-centric view of the world at any given time. \item \textbf{Scenario}: A scenario represents a time series of scenes, we use a temporal attribute, \texttt{timePosition} to differentiate the object behaviors across frames. 
\item \textbf{T-box}: Defines the overall structure of the ontology, inheriting the real-world assertions of dynamic object interactions in an ADS, serving as the ``blueprint". 
\item \textbf{A-box}: Contains the observed assertions from input images or frames. This includes the specific instances (e.g., \texttt{Vehicle\_1}, \texttt{Pedestrian\_1}), and their descriptive (e.g., \texttt{has\_color=Red}), and spatial  (e.g., \texttt{Vehicle\_1, is\_left\_of, Pedestrian\_1}) behavior. \end{itemize}

\subsection{Formalization}
Once the ontology graph captures the required core and dependent concepts to represent the real-world scenes consistently using common terminologies. The formalization allows curating complex behaviors (as shown in Table II) for identifying CP. This makes it possible to effectively differentiate scenes  e.g., finding occluded objects vs. missing objects using rules from HITL perspective. Rules are defined by universal quantification over properties (aka axioms), and combining them into queryable rules as shown in Table~\ref{tab:axioms_swr_rules}. While these rules are generated manually, it is possible to generate them through prompt engineering using LLMs too. \cite{lawrence_knowledge_graph_2023, ontoGPT, whetzel2011bioportal}.
\subsection{Reasoning}
The reasoning engine gains inferential power from CP formalization, by iteratively validating all possible combinations of asserted and inferred knowledge extracted from the KGs. This process uncovers implicit relationships and unique contextual patterns that are otherwise not derivable from the raw input data, e.g., being able to detect overlapping regions or conflicting spatial relationships in objects of interest through CP validation. Additionally, the reasoner helps in checking for logical inconsistencies in the graph, including missing attributes, contradictory relationships, and domain-specific rule violations.

We leverage third-party tools such as the Pellet reasoner \cite{Pellet}, the HermiT reasoner \cite{hermit2014}, and the Protégé ontology editor \cite{protege} to support reasoning. Pellet and HermiT provide description logic reasoning capabilities for classification, constraint validation, and entailment validation. Protégé is a user-friendly tool for ontology visualization, rule creation, and query prototyping.

\subsection{Validator and Feature Modification}

The validator module facilitates controlled experimentation by allowing the modification of CP specific node features. This method supports counterfactual analysis, addressing "what-if" scenarios to establish causal relationships. Currently, the module only supports two operations:
\begin{itemize} 
    \item Modifying pixel values around a misclassified object can help determine whether the failure is due to visual attributes like color or texture. 
    \item Modifying the scale of a segment can help determine whether errors are related to the size or proportion of objects within the system.
\end{itemize}

\subsection{Human-in-the-Loop}
To enhance accuracy and reliability, a HITL approach is integrated into various stages of the pipeline. This includes: 
\begin{itemize} 
    \item Validating segmentation outputs to ensure accuracy, 
    \item Refining formalization rules to align with domain-specific knowledge, 
    \item Overseeing the reasoning process to resolve ambiguities. 
\end{itemize}

Human oversight is valuable in complex or ambiguous cases, ensuring that the system’s outputs remain meaningful and actionable. This collaborative approach between automated systems and human experts enhances the robustness and applicability of the pipeline.

\section{Experiments \& Results}

This section demonstrates the capabilities of the CAIRO framework in identifying CP and analyzing potential root causes of perception failures in autonomous systems. Using two representative datasets, we showcase the framework’s ability to process diverse real-world scenarios and provide actionable insights. While these demonstrations highlight CAIRO’s potential, they are not formal evaluations with quantitative metrics.

\subsection{Dataset}
For our demonstrations, we utilized two datasets:
(1) CityScape~\cite{cityscape}, a well-known dataset featuring annotated urban traffic scenes, which serves as a benchmark for testing perception systems in structured environments. (2) Driving Downtown - New York City 4K~\cite{ny_video}, a video capturing real-world traffic interactions in dynamic urban conditions.

\begin{figure}[!h]
\centering
\includegraphics[width=\linewidth]{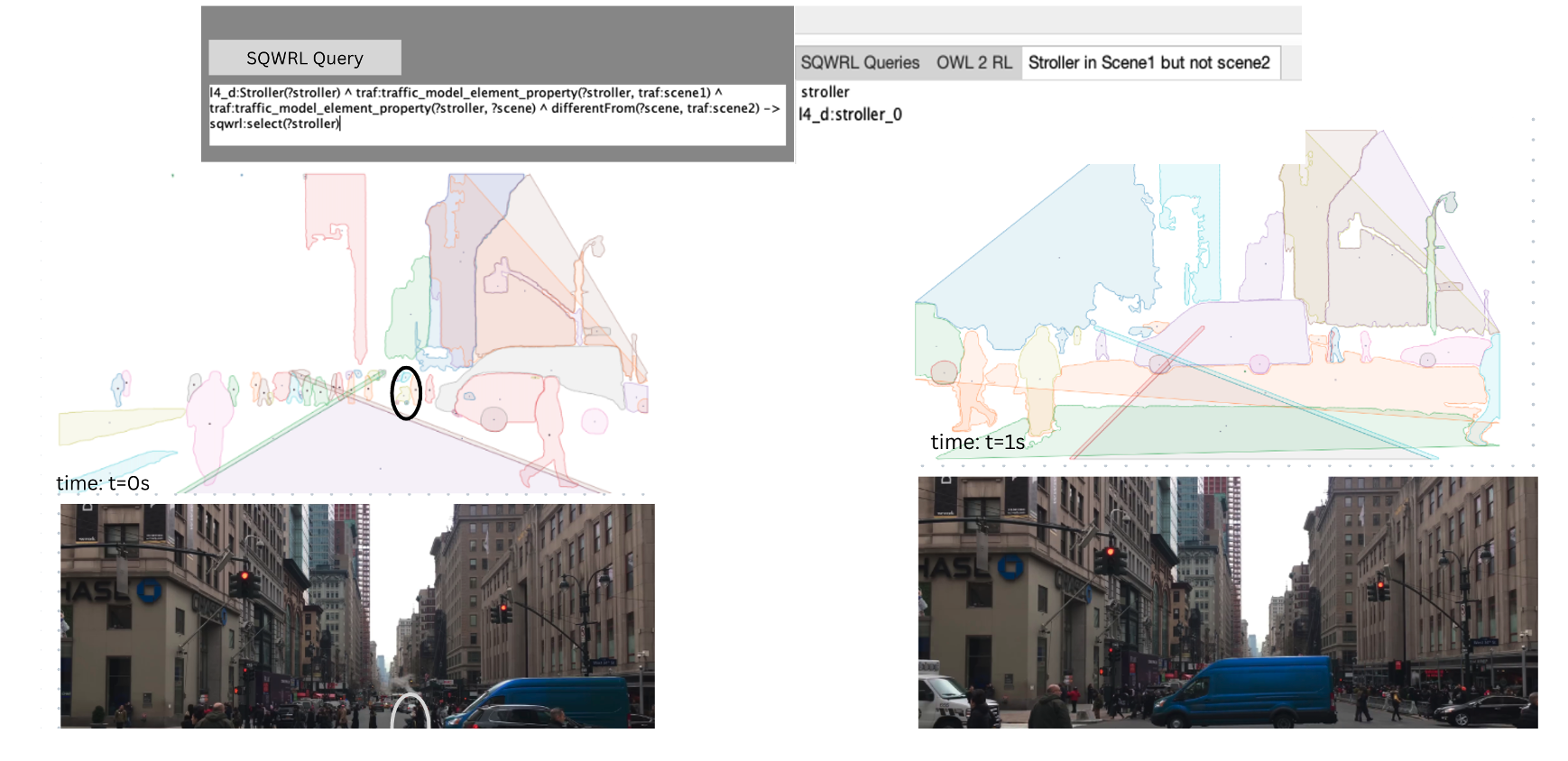}
\caption{\texttt{CP\_0002} from Table \ref{tab:cp_queries_combined} showing temporal relations, highlighting objects (e.g., stroller) that appears in one time frame but missing in the
consequent one.
}
\label{stroller}
\end{figure}

\subsection{Experimental Setup}

The experiments were conducted using a Tesla P100 GPU with 12 GB of memory. The results and outputs are available in our CAIRO\_Artifacts.
We utilized Pellet~\cite{Pellet} as the reasoner to ensure for logical consistency of the DM generated OWL scenarios.
In our experiments, we also included realistic universal patch attacks and noise-based adversarial attacks, such as those generated by optimized Stable Diffusion, which can alter backgrounds as potential criticality cases. However, new CP are added only when scoped concepts show discrepancies. Since CP are finite and are curated to highlight failure cases of a context, the same CP can be applied across many different images, ensuring consistency and efficiency in our analysis.

\subsection{Experiment 1: General CP}
To evaluate CAIRO’s versatility, we analyze CP involving behavioral unpredictabilities/dynamic object properties/DM artefacts. Figure \ref{other_cps} illustrates CP of a highly occluded road user, a detached wheel artefact, and a bicyclist illegitimately using the pedestrian crosswalk.

The experiments evaluates the CAIRO pipeline's ability to identify CP in safety-critical scenarios. Table \ref{tab:cp_queries_combined} outlines examples of CP using natural language and corresponding Semantic Query-Enhanced Web Rule Language (SQWRL) queries, focusing on key properties, such as dynamic object properties, distance metrics, behavioral unpredictability, and temporal relations. These CP address criticalities such as vehicle occlusion, proximity to driving lanes, and absence of attributes (i.e., license plate) and flags potential risks in diverse environments from a busy urban crosswalk to a near-vacant desert road.

Figure~\ref{stroller} illustrates CAIRO’s reasoning process for a CP where a stroller is detected near a drivable lane in time 0 but not in time 1. By leveraging SQWRL queries, the reasoner highlights temporal relationships by flagging the stroller’s temporary absence between consecutive time frames, demonstrating its ability to process temporal object relations.

\begin{figure}[h]
\centering
\includegraphics[width=\linewidth]{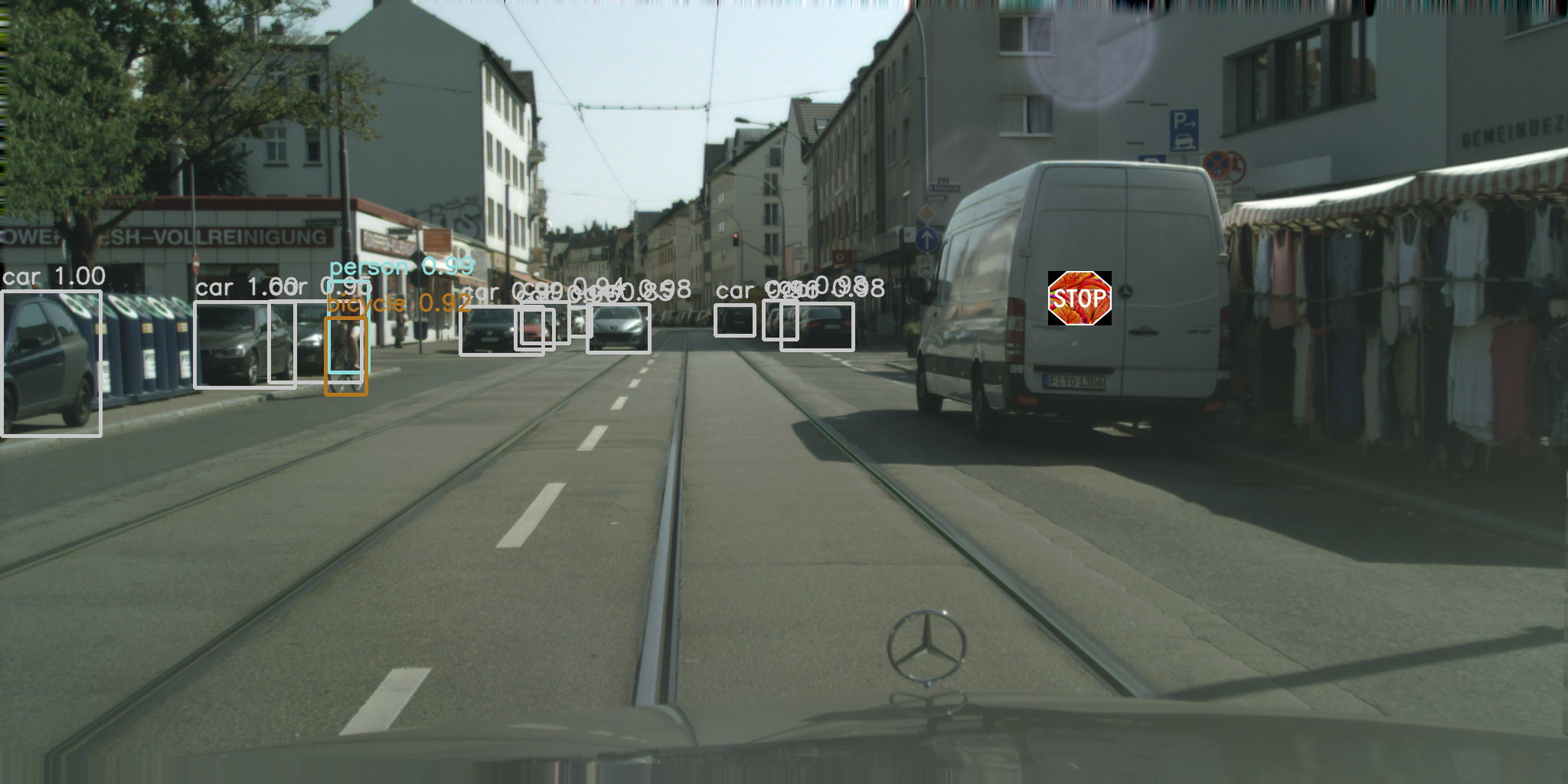}
\caption{Patch Attack from ShapeShifter~\cite{ShapeShifter} and placed in a way to fool Faster R-CNN which completely missed the vehicle}
\label{patch_attack}
\end{figure}

\begin{figure}[h]
\centering
\includegraphics[width=\linewidth]{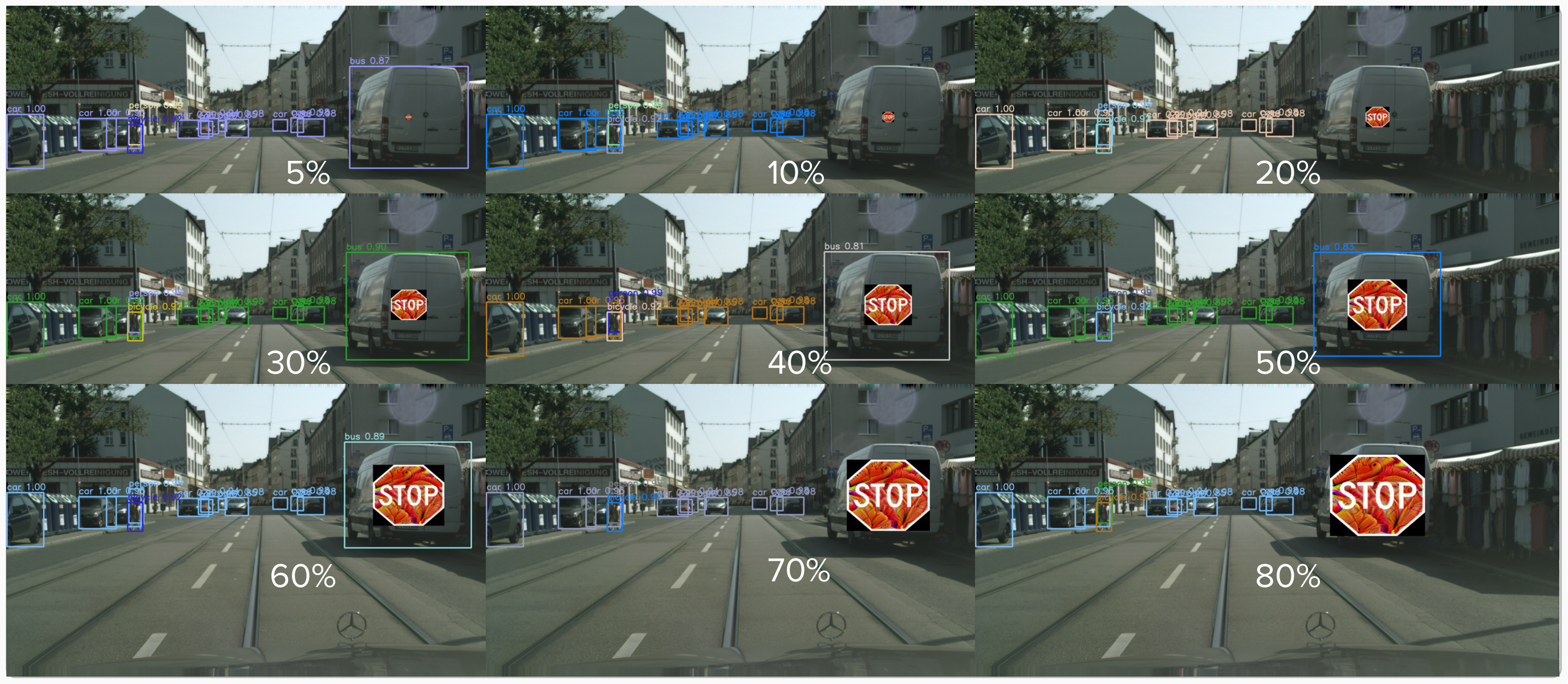}
\caption{Impact of Modifying Segment Occlusion Rate from 5\% to 80\% }
\label{patch_scale}
\end{figure}

\subsection{Experiment 2: Adversarial CP}

In this experiment, we selected a single frame, \texttt{frankfurt\_000000\_001751.png}, from the CityScape dataset~\cite{cityscape} for a patch attack. Figure ~\ref{patch_attack} illustrates that Faster R-CNN completely missed the large vehicle due to a small stop sign patch. Here are the steps detailing how the CAIRO pipeline operates:

\begin{enumerate}
    \item CAIRO constructor extracted all necessary segments from three layers of segmentation. The adversarial patch itself was still classified by CLIP as a traffic sign in the ontology graph.
    \item CAIRO reasoner triggered a CP (Traffic Sign On Vehicle): 
    \texttt{$\textit{TrafficSign}(ts) \land \textit{is\_part\_of}(ts, v) \land \textit{Vehicle}(v) \land \textit{highOcclusion}(v) \rightarrow \textit{CP}(ts)$
    }, where \textit{highOcclusion} is a Boolean value, indicating whether the occlusion rate exceeds pre-defined thresholds.

    \item The property of the segment/node \texttt{traffic\_post} was modified, changing the RGB values and the scale of the patch. 
    
    \item Figure \ref{patch_scale} shows the results from the validation process, which adjusted the scale of the CP segment to a specific ratio. When the occlusion rate ( x ) is either (x $\leq$ 5\%) or (30\% $\leq$ x $\leq$ 60\%), Faster R-CNN is able to detect the vehicle. 
\end{enumerate}

\begin{table}[h]
    \centering
    \caption{Average Elapsed Time per Frame (\textbf{F}: Faster R-CNN, \textbf{G}: GroundingDINO, \textbf{C}: CLIP \textbf{L}: Lane Detector)}
    \begin{tabular}{|l|c|c|c|c|c|}
        \hline
        & \textbf{F} & \textbf{G} & \textbf{C} & \textbf{L} & \textbf{Pellet Reasoner} \\ \hline
        \textbf{Per Frame (s)} & 1.69 & 2.14 & 146.51 & 0.14 & 865.37 \\ \hline
    \end{tabular}
    \label{tab:extraction_times}
\end{table}

Table~\ref{tab:extraction_times} shows the average elapsed time for each frame from construction to reasoning. The current CAIRO system lacks a module for providing natural language explanations for Faster R-CNN failure reasons. However, specific reasons can be inferred. The system's ability to re-detect the vehicle by modifying the adversarial patch scale suggests potential limitations in the pre-training dataset related to a lack of diversity in both object scale and context. Factors contributing to these failures may include but are not limited to the following: 1) Scale Sensitivity: the model may only recognize stop signs within specific size ranges due to limited examples in the training data. 2) Contextual Understanding: the model might heavily rely on typical object sizes and surroundings, creating confusion when encountering unusual sizes. 3) Data Diversity: the training data might not be versatile enough, leading to less model adaptability.

\begin{figure}[h]
\centering
\begin{minipage}{0.32\columnwidth}
    \centering
    \includegraphics[width=\linewidth]{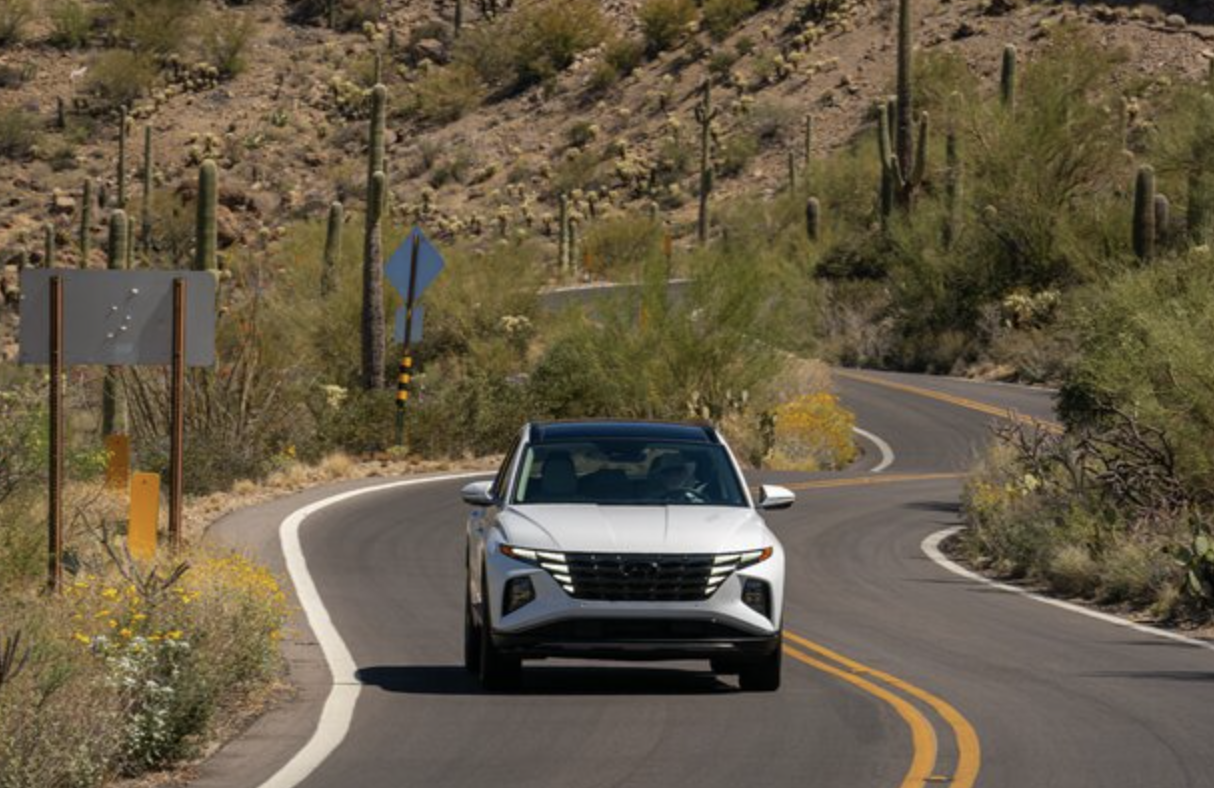}
    \small{(a) Original Image}  
\end{minipage}
\hfill
\begin{minipage}{0.33\columnwidth}
    \centering
    \includegraphics[width=\linewidth]{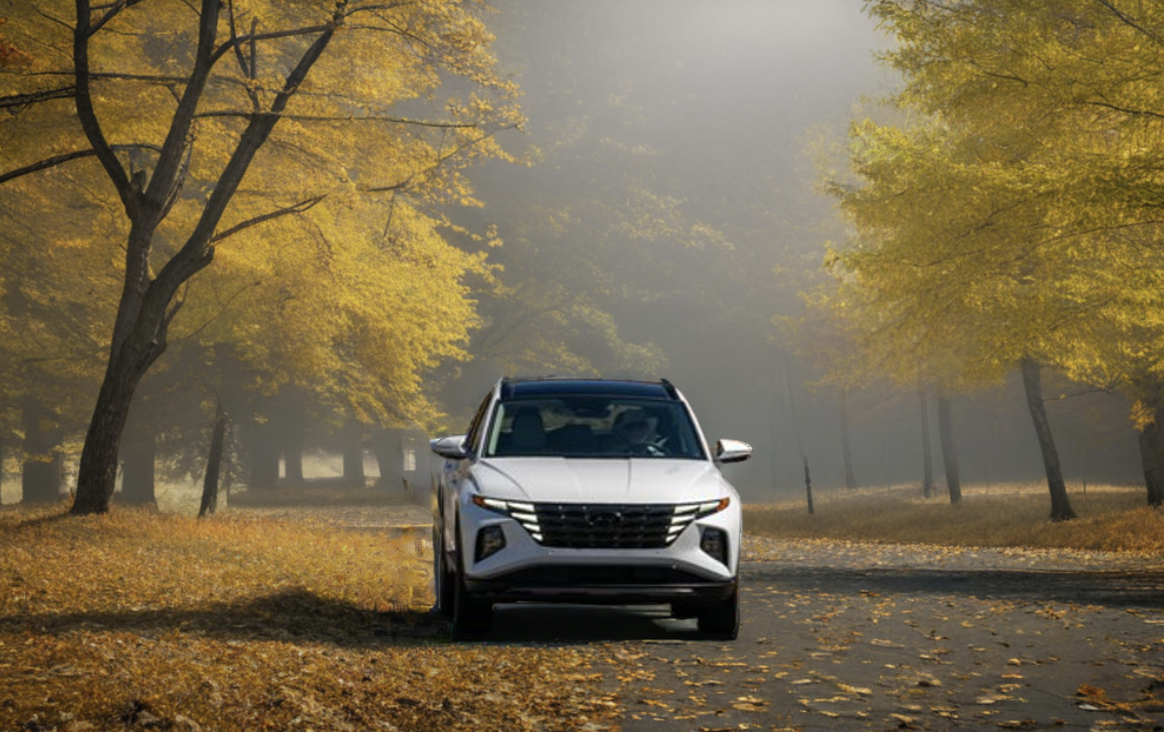}
    \small{ (b) Generated Image}  
\end{minipage}
\hfill
\begin{minipage}{0.32\columnwidth}
    \centering
    \includegraphics[width=\linewidth]{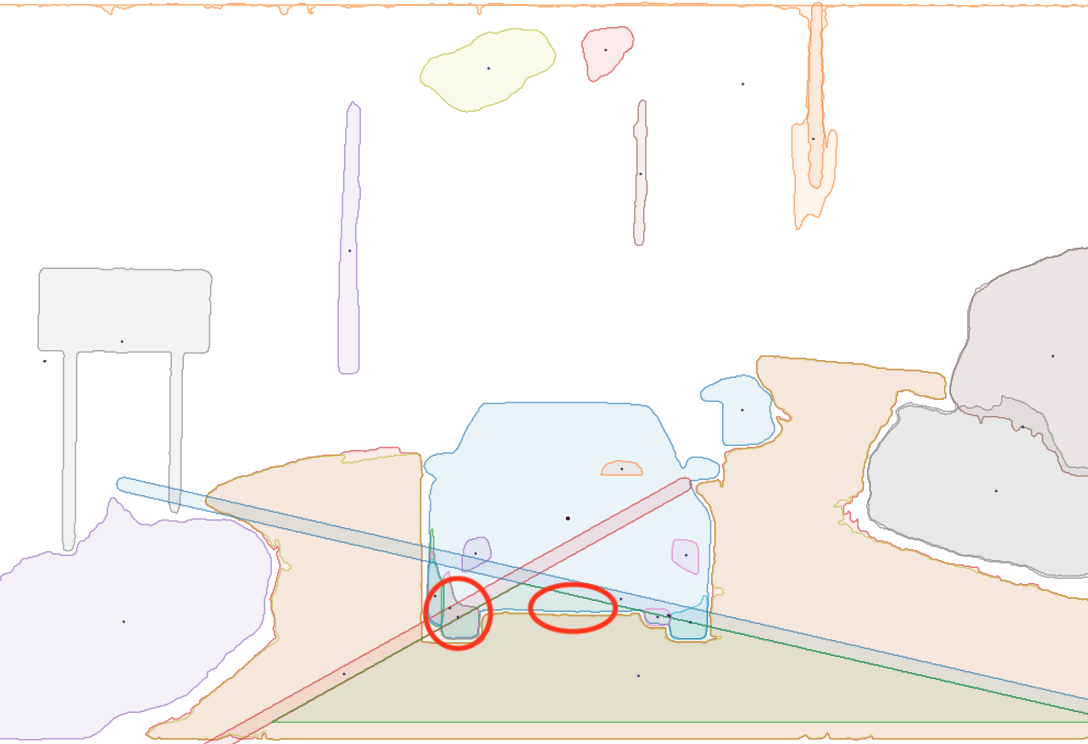}
    \small {(c) Common CP}  
\end{minipage}
\caption{\small Comparison of the original and generated images, along with the common CP that is shared by both.}
\label{fig:comparison}
\end{figure}

\subsection{Experiment 3: Noise Optimization Attack}

We started by testing the reasoning system on an original image of a car driving on a desert road (see Figure \ref{fig:comparison}a). When passed through the Faster R-CNN model (COCO pre-trained), the car was detected with 97\% confidence. The CAIRO reasoner flagged two specific CP: \texttt{car with disproportionate wheels} and \texttt{missing license plate}. This demonstrates that the reasoner can detect object-level issues, such as a car’s wheel size, or the presence of a license plate even in a clear and simple scene.

Next, we modified the image using the Stable Diffusion inpainting model by changing the background of the object of interest (e.g., a car) to challenge the DM (see Figure \ref{fig:comparison}b). The denoising process, driven by the model’s loss function, preserved the car but replaced clear road markers and boundaries with trees and foliage. In the generated image, the DM misdetected a car as a person with a confidence of 99\%. Despite this error, the reasoning system still detected the original CP, showing it can reliably track object-level problems across different scenarios.

In addition to the aforementioned two CP — \texttt{disproportionate wheels} and \texttt{missing license plate} (see Figure \ref{fig:comparison}c)— the system also identified a new CP: \texttt{missing lane markers}. By replacing predefined road features for more natural elements, the scene became ambiguous, highlighting an important point: while the reasoning system is effective at detecting object-level issues, significant changes to the scene can introduce new uncertainties, reducing the detection model’s confidence.

\section{Conclusion}

This paper introduces CAIRO, a novel framework that
combines ontological knowledge graphs and HITL validation for an accountable T\&E of critical systems, particularly in AV applications. Using contextual reasoning and structured domain knowledge, CAIRO addresses the challenges of detecting and interpreting CP in the input scenes, including adversarial attacks, behavioral anomalies, and perception artefacts in real-world scenarios. In our approach we share our domain-specific knowledge building process for AV from known publications, authoritative sources, and open-source tools. Our experiments demonstrate the framework’s capacity to identify root causes of perception failures and enhance object detection reliability. CAIRO’s reasoning capabilities uncover insights that bridge the gap between AV perception and human cognition, promoting safer and more interpretable AV systems. The framework not only provides actionable recommendations for system improvement but also introduces scalable methods for sharing and extending ontological models across diverse applications. Future work will explore CAIRO’s potential in mitigating evolving real-world threats in different domains with collaborative efforts in human-centered AI.

\bibliographystyle{unsrt}
\bibliography{main}

\begin{IEEEbiographynophoto}{Sridevi Polavaram}
Sridevi Polavaram received her BS (applied mathematics and electronics) and MS (computer applications) from Osmania University in Hyderabad, India. She later moved to the USA and pursued a second master’s in computer science, which was followed by a PhD in neuroscience from George Mason University in Fairfax, Virginia in 2016.  Her dissertation focused on neuroinformatic tools for digitization and feature profiling on shared data (of 3D neuronal reconstructions) from diverse experimental sources. After receiving her doctorate, Dr. Polavaram gained experience as a data scientist for CareFirst BlueCross BlueShield before accepting a Senior Data Scientist position at the MITRE Corporation in McLean, Virginia. At MITRE, she currently works as the Lead AI Researcher in projects regarding the development of responsible AI models for a variety of uses, from health analytics and biometrics to law enforcement and human-machine teaming.
\end{IEEEbiographynophoto}

\begin{IEEEbiographynophoto}{Meenu Ravi}
received a BS in Computer Science from Rensselaer Polytechnic Institute. She later pursued her MS in Analytics from the University of Chicago and is currently pursuing her PhD in Computer Science at Virginia Tech. Since 2021, she has been a data scientist working in aviation, health anlytics, and economic tariff fields at MITRE Corporation in McLean, VA.
\end{IEEEbiographynophoto}

\end{document}